\documentclass[10pt,twocolumn,letterpaper]{article}

\usepackage{cvpr}              

\usepackage{graphicx}
\usepackage{amsmath}
\usepackage{amssymb}
\usepackage{booktabs}

\usepackage[pagebackref,breaklinks,colorlinks]{hyperref}

\makeatletter
\renewcommand{\maketag@@@}[1]{\hbox{\m@th\normalsize\normalfont#1}}%
\makeatother

\usepackage[capitalize]{cleveref}
\crefname{section}{Sec.}{Secs.}
\Crefname{section}{Section}{Sections}
\Crefname{table}{Table}{Tables}
\crefname{table}{Tab.}{Tabs.}

\begin{document}

\title{Semi-Supervised 2D Human Pose Estimation Driven by Position Inconsistency Pseudo Label Correction Module}


\author{
Linzhi Huang\textsuperscript{1, 2} \thanks{This work was done when the authors were visiting Beike as interns.} \\
\and Yulong Li\textsuperscript{2} \\
\and Hongbo Tian\textsuperscript{1, 2*} \\
\and Yue Yang\textsuperscript{2} \\
\and Xiangang Li\textsuperscript{2} \\
\and Weihong Deng\textsuperscript{1 \thanks{Corresponding author.}}  \\
\and Jieping Ye\textsuperscript{2} \\
\and {\textsuperscript{1}Beijing University of Posts and Telecommunications, \textsuperscript{2}Beike} \\
 {\tt\small\{huanglinzhi, tianhongbo, whdeng\}@bupt.edu.cn} \\
 {\tt\small\{liyulong008, yangyue092, lixiangang002, yejieping\}@ke.com}
}

\maketitle

\begin{abstract}
In this paper, we delve into semi-supervised 2D human pose estimation.
The previous method ignored two problems: 
(i) When conducting interactive training between large model and lightweight model, the pseudo label of lightweight model will be used to guide large models.
(ii) The negative impact of noise pseudo labels on training.
Moreover, the labels used for 2D human pose estimation are relatively complex: keypoint category and keypoint position.
To solve the problems mentioned above, we propose a semi-supervised 2D human pose estimation framework driven by a position inconsistency pseudo label correction module (SSPCM).
We introduce an additional auxiliary teacher and use the pseudo labels generated by the two teacher model in different periods to calculate the inconsistency score and remove outliers.
Then, the two teacher models are updated through interactive training, and the student model is updated using the pseudo labels generated by two teachers.
To further improve the performance of the student model, we use the semi-supervised Cut-Occlude based on pseudo keypoint perception to generate more hard and effective samples.
In addition, we also proposed a new indoor overhead fisheye human keypoint dataset WEPDTOF-Pose.
Extensive experiments demonstrate that our method outperforms the previous best semi-supervised 2D human pose estimation method.
We will release the code and dataset at https://github.com/hlz0606/SSPCM.
\end{abstract}

\begin{figure}[t]
\centering
\includegraphics[width=0.95\linewidth]{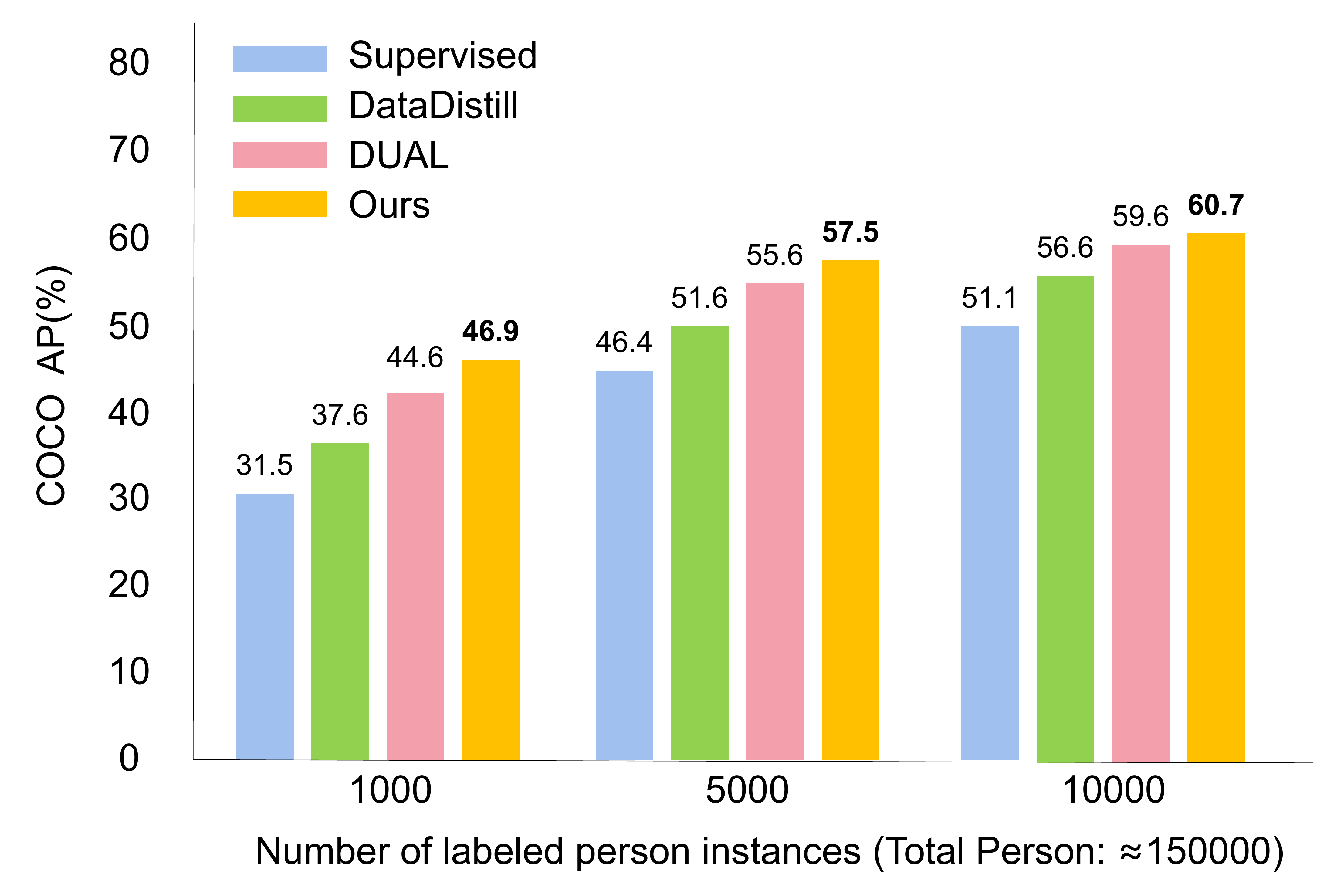}
\caption{
\textbf{Performance comparison} between our method SSPCM and SOTA method (DataDistill \cite{radosavovic2018data}, DUAL \cite{xie2021empirical}) on COCO \cite{lin2014microsoft} dataset.
On the COCO dataset, using 1000, 5000, and 10000 labeled person instances, our method has increased 2.3mAP, 1.9mAP, and 1.1mAP compared with the previous method.
}
\label{fig:performance_comparison}
\end{figure}

\begin{figure*}[t]
\centering
\includegraphics[width=0.95\linewidth]{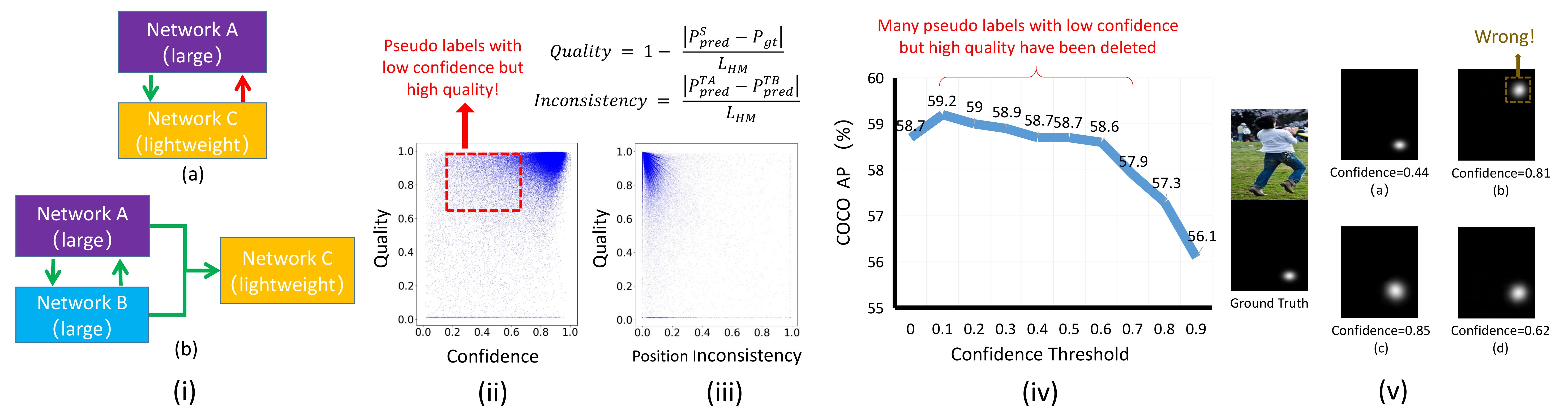}
\caption{
\textbf{Illustration of motivation.} 
(i): (a) Previous interactive training methods \cite{xie2021empirical}.
(b) Our architecture. 
The arrow indicates the transmission direction of the pseudo label.
(ii) and (iii) are the statistics of teacher model predictions (pseudo labels) on COCO dataset.
$P_{pred}$ represents the keypoint coordinates predicted by the model (based on the heatmap).
$S$ represents student model.
$TA$ and $TB$ represent 2 teacher models.
$P_{gt}$ represents the ground truth keypoint coordinates (based on the heatmap).
$L_{HM}$ represents the diagonal length of the heatmap.
\textbf{(ii)}: The relationship between the quality of the pseudo labels and confidence. 
\textbf{(iii)}: The relationship between the quality of the pseudo labels and position inconsistency.
\textbf{(iv)}: Results of the DUAL \cite{xie2021empirical} at different confidence thresholds.
\textbf{(v)}: One specific example to demonstrate that the confidence does not represent the localization quality.
(a) and (b) are the output results of teacher model A in different epochs (heatmap of the right ankle). 
(c) and (d) are the output results of the additional auxiliary teacher model B.
}
\label{fig:motivation}
\end{figure*}

\section{Introduction}
\label{sec:Introduction}
2D human pose estimation (HPE) \cite{bin2020adversarial, chen2018cascaded, li2021human, zhang2019fast, li2021online, newell2017associative} is a task to estimate all 2D keypoints of the human body from images. 
It is a fundamental task of action recognition \cite{duan2022revisiting, cai2021jolo, yan2018spatial}, 3D human pose estimation \cite{huang2022dh, pavllo20193d, zheng20213d, li2022mhformer, martinez2017simple}, etc.
In recent years, thanks to the development of deep learning \cite{hu2018squeeze, krizhevsky2012imagenet, simonyan2014very, szegedy2015going}, 2D human pose estimation has made significant progress. 
However, the training of such a task is known to be data-hungry, where the labelling process is particularly costly and time-consuming.
To solve this problem, semi-supervised 2D human pose estimation has become an important research direction. 
This direction focuses on how to use a small amount of labeled data and a large amount of unlabeled data to improve the performance of the model.

The current state-of-the-art semi-supervised 2D HPE model \cite{xie2021empirical} is based on consistency learning.
Xie et al. \cite{xie2021empirical} find that by maximizing the similarity between different increments of the image directly, there would be a collapsing problem.
The reason is that the decision boundary passes the high-density areas of the minor class, so more and more pixels are gradually wrongly classified as backgrounds.
They proposed a simple way to solve this problem. 
For each unlabeled image, an easy augmentation $I_{e}$ and a hard augmentation $I_{h}$ are generated, and they are fed to the network to obtain two heatmap predictions. 
They use the accurate predictions on the easy augmentation to teach the network to learn about the corresponding hard augmentation.
In addition, they also proposed a method that can replace EMA \cite{grill2020byol} to update the parameters of the teacher model, called Dual Network.
The two models will take turns to act as teachers' identities to generate pseudo labels, and take turns to act as students' identities.

The previous methods \cite{radosavovic2018data, xie2021empirical} can improve the accuracy of student models.
However, the previous method has the following problems:
1) In the practical application of semi-supervised learning, large model are often used as teacher and lightweight model as student.
Due to the inconsistent model structure, it is hard to use EMA to update the teacher model.
When conducting interactive training between large model and lightweight model, the pseudo label of lightweight model will be used to guide large models, as shown in the (a) of Fig. \ref{fig:motivation} (i).
Although this method can also improve the performance of the teacher model, it is suboptimal.
2) The noise labels will harm the model training, and the student model will overfit the noise labels (causing confirmation bias \cite{arazo2020pseudo}).
Some previous semi-supervised classification tasks \cite{sohn2020fixmatch, lee2013pseudo} use the confidence of classification to filter pseudo labels.
There are a large number of high-quality pseudo labels in the low confidence region, as shown in Fig. \ref{fig:motivation} (ii).
When the confidence threshold exceeds a certain value, the higher the confidence threshold, the lower the model performance, as shown in the Fig. \ref{fig:motivation} (iv).
By observing (a) and (b) in Fig. \ref{fig:motivation} (v), we can find that (b) has a higher confidence, but it is a noise label deviating from the ground truth (outliers).
Therefore, we choose to filter with a lower threshold.
In addition, in recent studies\cite{wang2021data, yang2022mix, rizve2021defense, li2022pseco, jiang2018acquisition}, it has been found that the more inconsistent the prediction results of different models for the same object, the more likely the prediction results will be wrong.
To solve above problems, we introduce an additional auxiliary teacher and use it to generate pseudo labels. 
Their parameters are updated through interactive training, which ensures the difference between the two models, as shown in the (b) of Fig. \ref{fig:motivation} (i).
The structure of these two teacher models can be consistent with the student model, or they can be larger models.
Two teacher models may output different results for the same image.
Even in different training periods, the output results of the same model for the same image will be different.
We post-process the $N$ pseudo labels output by the two teacher models in different periods to obtain $N$ prediction results (2D coordinates) of each keypoint.
Then, we calculate the pixel distance between $N$ prediction results of each keypoint.
We use pixel distance to characterize the degree of position inconsistency (inconsistency score).
We visual the relationship between the quality of the pseudo labels and position inconsistency, as shown in Fig. \ref{fig:motivation} (iii).
We select a group of pseudo labels (2 pseudo labels) with the smallest position inconsistency for ensemble to obtain the final corrected pseudo labels.
In short, on the basis of filtering based on confidence, PCM module selects a set of pseudo labels with the least inconsistency to remove outliers.
The correction of pseudo labels of PCM module is similar to ensemble learning, which can make pseudo labels smoother.
It is worth mentioning that we only use student model when testing.

In addition, we also use the semi-supervised Cut-Occlude based on pseudo keypoint perception to generate more hard samples, as shown in Fig. \ref{fig:my_augmentation}.
Specifically, we use the pseudo label of the teacher model to locate the center of each keypoint in the image.
Then, based on this central position, we cut out the local limb image.
We randomly paste the local limb image to the center of a keypoint in another image to simulate local occlusion.


Our contributions are as follows:
\begin{itemize}
	\item 
        We propose a semi-supervised 2D human pose estimation framework driven by a position inconsistency pseudo label correction module (SSPCM).
        Especially when the structure of teacher model and student model is inconsistent, it is a better solution.

	\item 
	    To further improve the performance of the student model, we propose the semi-supervised Cut-Occlude based on pseudo keypoint perception (SSCO) to generate more hard and effective samples.
     
	\item
	    Extensive experiments on MPII \cite{andriluka14cvpr}, COCO \cite{lin2014microsoft}, and AI-Challenger \cite{wu2019large} have proved that our method outperforms the previous best semi-supervised 2D human pose estimation method
     , as shown in Fig. \ref{fig:performance_comparison}.
	\item
	    We release a new 2D HPE dataset collected by indoor overhead fisheye camera based on the WEPDTOF \cite{tezcan2022wepdtof} dataset, which is called WEPDTOF-Pose.
	    We have conducted lots of experiments on WEPDTOF-Pose, CEPDOF \cite{duan2020rapid} and BKFisheye datasets (after removing sensitive information).
	    
\end{itemize}


\begin{figure*}[t]
\centering
\includegraphics[width=0.95\linewidth]{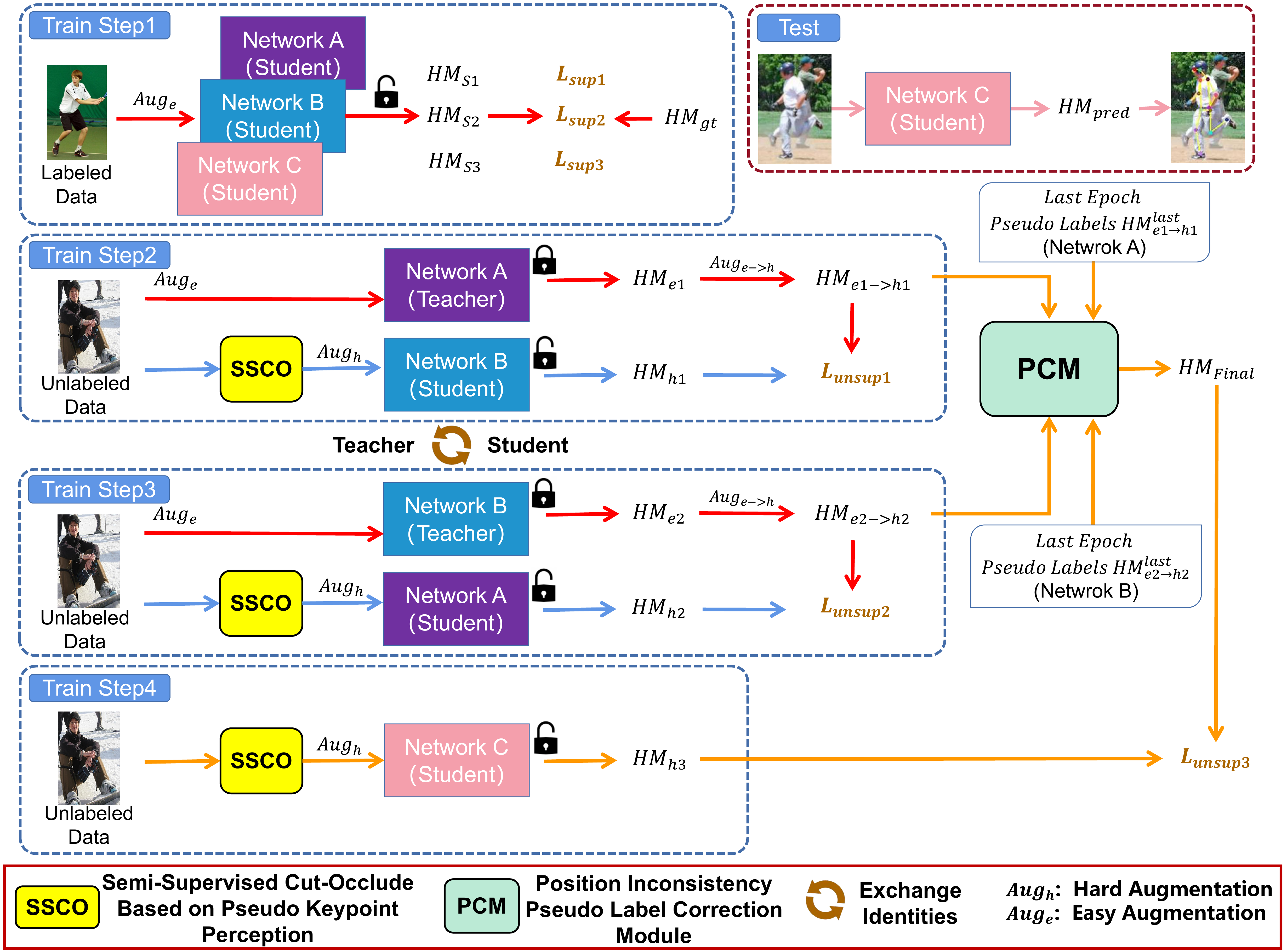}
\caption{
\textbf{Overall overview of our SSPCM.}
In Train Step 1, we use labeled data to train $Network A$, $Network B$, and $Network C$, and update their parameters.
In Train Step 2 and Train Step 3, we use unlabeled data and interactive training to update the parameters of $Network A$ and $Network B$.
In Train Step 4, we use $Network A$, $Network B$, and PCM modules to guide $Network C$ training. 
SSCO is the semi-supervised Cut-Occlude based on pseudo keypoint perception. 
When testing, we only use $Network C$.
}
\label{fig:architecture}
\end{figure*}

\section{Related Work}
\label{sec:Related Work}

\textbf{2D human pose estimation.} 
2D human pose estimation (HPE) \cite{bin2020adversarial, chen2018cascaded, li2021human, zhang2019fast, li2021online, newell2017associative} is one of the most important tasks in computer vision.
Its purpose is to detect the keypoints of the human body from the image and predict the correct category.
2D HPE can generally be divided into two methods: top-down and bottom-up.
The top-down method divides the whole task into two stages: human detection and keypoint detection.
To be specific, we first use human detection to obtain human bbox, and then use human pose estimation to obtain the keypoints of each human.
For example, HRNet \cite{sun2019hrnet} proposes a multi-scale feature fusion structure, which maintains a high-resolution representation and can achieve very good results on COCO \cite{lin2014microsoft} and other datasets.
The bottom-up method is to first detect all the keypoints in the original image, and then assign these keypoints to the corresponding human body.
For example, HigherHRNet \cite{cheng2020higherhrnet} proposes to use high-resolution feature pyramids to obtain multi-scale information and uses association embedding \cite{newell2017associative} to group keypoints.
However, 2D HPE needs to label the keypoints of each human body in the dataset, which is labor-intensive and expensive.
Therefore, we propose a new semi-supervised 2D human pose estimation framework to mitigate this problem.

\textbf{Semi supervised learning (SSL).}
Semi-supervised learning uses a small amount of labeled data and a large amount of unlabeled data to train the model.
The current semi-supervised methods are mainly divided into SSL based on the pseudo label \cite{lee2013pseudo, radosavovic2018data, xie2020self, yarowsky1995unsupervised} and SSL based on consistency \cite{sajjadi2016ts,laine2017temporal,tarvainen2017mean,berthelot2019mixmatch,sohn2020fixmatch}.
SSL based on pseudo labels generates pseudo labels for unlabeled data through pretrained models and uses these pseudo labels to further optimize the model.
Consistency-based SSL enables multiple images to be obtained by different data augmentation to the same image and encourages the model to make similar predictions about them.
For example, FixMatch \cite{sohn2020fixmatch} uses the model to generate pseudo labels for weakly augmented unlabeled images.
Only when the model produces a prediction with high confidence will the pseudo label be retained.
Then, when a strongly augmented version of the same image is input, the model is trained to predict the pseudo labels.
We mainly focus on SSL based on consistency, because it has superior accuracy in the public benchmark.

\textbf{Semi-supervised 2D human pose estimation.}
The goal of semi-supervised 2D human pose estimation is to optimize the performance of the human pose estimator using a small amount of labeled data and a large amount of unlabeled data.
Xie et al. \cite{xie2021empirical} find that by maximizing the similarity between different increments of the image directly, there would be a collapsing problem.
They propose a Dual \cite{xie2021empirical} network to solve this problem. 
First, the input image is augmented into a pair of hard and easy data, and the easy augmentation data is transferred to the teacher model and the hard augmentation data is transferred to the student model to keep the output of the two models consistent.
In addition, they also update the parameters by letting the two models take turns playing the roles of teachers and students, which is better than using EMA \cite{grill2020byol} directly.
However, they ignore the negative impact of noise pseudo labels on training.
Therefore, we propose a new semi-supervised training framework and a new data augmentation method.

\section{Method}
\label{sec:Method}

In this section, we first give the definition of the semi-supervised 2D human pose estimation task (see Sec. \ref{sec:Problem_Definition}). 
Then, in Sec. \ref{sec:PCM}, we introduced a semi-supervised 2D human pose estimation framework based on the position inconsistency pseudo label correction module. 
Finally, we introduced the semi-supervised Cut-Occlude based on pseudo keypoint perception in Sec. \ref{sec:SSCO}.

\subsection{Problem Definition}
\label{sec:Problem_Definition}
In semi-supervised 2D human pose estimation (SSHPE), we obtained a set of labeled data $D_{l}={\{(x_{i}^{l}, y_{i}^{l})\}}_{i=0}^{n_{l}}$ and a set of unlabeled data $D_{u}={\{(x_{j}^{u})\}}_{j=0}^{n_{u}}$, where $x$ and $y$ represent images and ground truth labels, $n_{l}$ represents the number of labeled data, and $n_{u}$ represents the number of unlabeled data.
The goal of SSHPE is to train 2D human pose estimators on labeled and unlabeled data.
The loss function is as follows:
\begin{equation}
    L_{all}=\sum_{i}{L(x_{i}^{l}, y_{i}^{l})} + \gamma\cdot\sum_{j}{L(x_{j}^{u}, y_{j}^{u})}
\end{equation}
where $x_{i}^{l}$ represents labeled data, $y_{i}^{l}$ represents ground truth label, $x_{j}^{u}$ represents unlabeled data, $y_{j}^{u}$ represents pseudo label generated by teacher model, $\gamma$ represents weight of unsupervised learning, and $L$ represents loss.

\begin{figure}[t]
\centering
\includegraphics[width=0.95\linewidth]{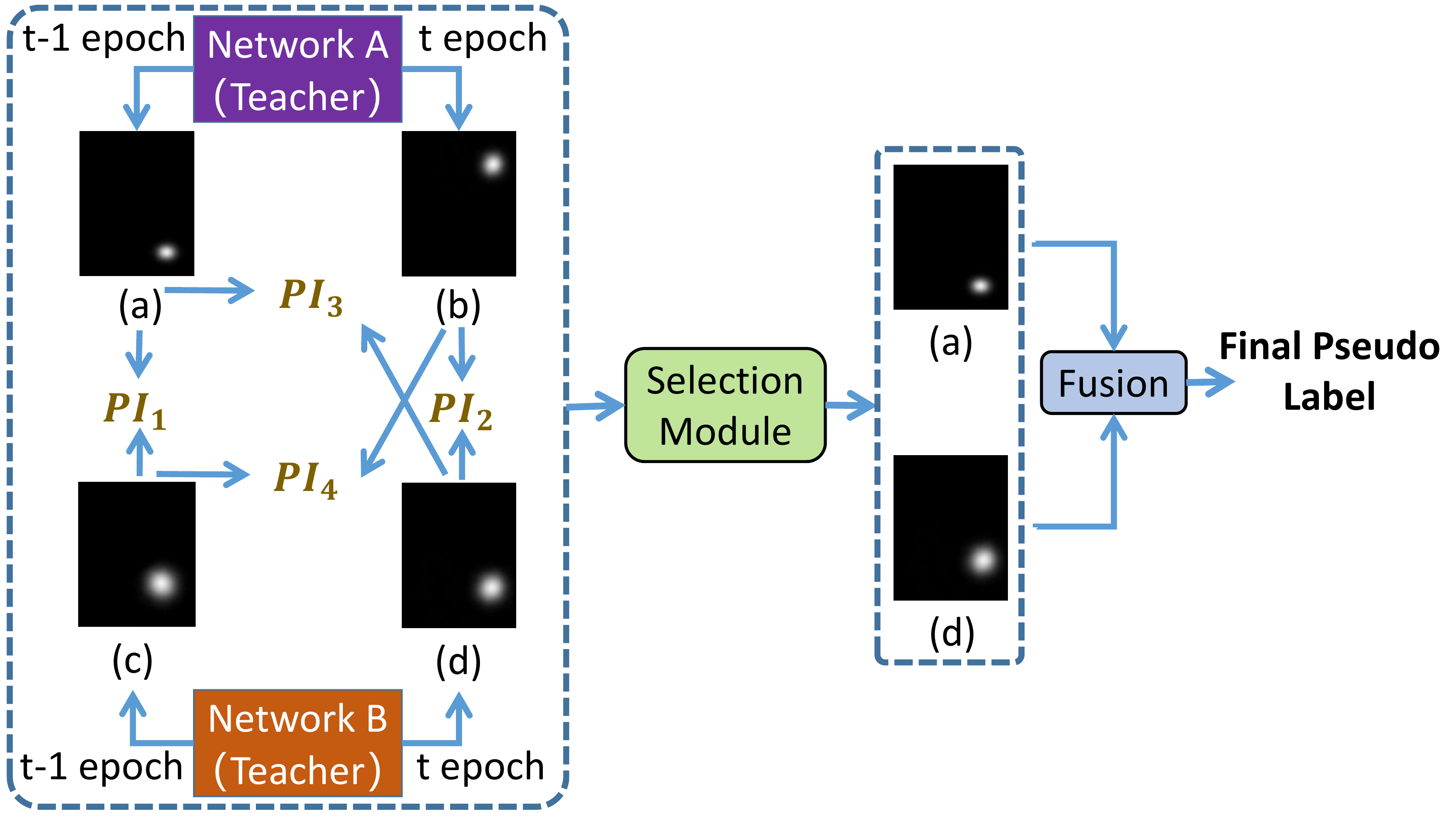}
\caption{
\textbf{Position inconsistency pseudo label correction module (PCM).}
(a) Pseudo label of $Network A$ output in the last epoch. 
(b) Pseudo label of $Network A$ output in the current epoch. 
(c) Pseudo label of $Network B$ output in the last epoch. 
(d) Pseudo label of $Network B$ output in the current epoch.
We post-process the pseudo labels to get the pseudo keypoint coordinates. 
Next, we calculate the position inconsistency $PI$ between pseudo keypoints output by different models and transfer it to the Selection Module.
Then, the Selection Module selects a group of pseudo labels with the smallest $PI$ and performs pseudo labels fusion to obtain the final corrected pseudo labels.
}
\label{fig:Fusion_method}
\end{figure}

\subsection{Overview of SSPCM}
\label{sec:PCM}
Fig. \ref{fig:architecture} shows the overall framework of our SSPCM. 
We will introduce SSPCM in detail in this section.
As described in Sec. \ref{sec:Introduction}, we introduced an auxiliary $Network B$ ($f_{\theta}^{B}$) on the basis of the original $Network A$ ($f_{\theta}^{A}$) and $Network C$ ($f_{\theta}^{C}$), where $\theta$ represents network parameters.
The three models have the same network structure, but their parameters are independent.
In training, the training process for each batch of data can be divided into 4 stages, as shown in Fig. \ref{fig:architecture}.
Next, we will introduce these 4 steps in detail. 
The PCM module will be introduced in Train Step 4.

\textbf{Train Step 1.}
$Network A$ ($f_{\theta}^{A}$), $Network B$ ($f_{\theta}^{B}$) and $Network C$ ($f_{\theta}^{C}$) trains on labeled data and updates parameters.  
The supervision losses are as follows:
\begin{tiny}
\begin{equation}
    L_{sup}=\sum_{n{\in}N}{{\|{HM_{gt}^{n} - HM_{s1}^{n} }\|}^{2}  + {\|{HM_{gt}^{n} - HM_{s2}^{n} }\|}^{2} + {\|{HM_{gt}^{n} - HM_{s3}^{n} }\|}^{2}}
\end{equation}
\end{tiny}
where $HM_{gt}^{n}$ represents the ground truth of the $n$th image in the labeled data.
$HM_{s1}^{n}$, $HM_{s2}^{n}$ and $HM_{s3}^{n}$ respectively represent the prediction results of $Network A$ ($f_{\theta}^{A}$), $Network B$ ($f_{\theta}^{B}$) and $Network C$ ($f_{\theta}^{C}$) on the $n$th image in the labeled data.

\textbf{Train Step 2.}
First, $Network A$ ($f_{\theta}^{A}$) is used as the teacher model (with fixed parameters), and the additional $Network B$ ($f_{\theta}^{B}$) is used as the student model (with parameter updates).
Next, easy data augmentation $Aug_{e}$ is performed on unlabeled data $I_{u}$, and the predicted pseudo labels $HM_{e1}$ are obtained by inputting them into $Network A$ ($f_{\theta}^{A}$).
Then, input unlabeled data $I_{u}$ into the SSCO module (detailed in Sec. \ref{sec:SSCO}) to get the hard sample with occlusion, and hard data augmentation $Aug_{h}$ is performed on it.
We input this hard sample into $Network B$ ($f_{\theta}^{B}$) to get the prediction results $HM_{h1}$.
Finally, we use $Aug_{e{\to}h}$ to map $HM_{e1}$ to $HM_{e1{\to}h1}$ and calculate the consistency loss between $HM_{e1{\to}h1}$ and $HM_{h1}$:
\begin{equation}
    L_{unsup1}=\sum_{n{\in}N}{{\|{HM_{e1{\to}h1}^{n} - HM_{h1}^{n} }\|}^{2}}
\end{equation}
where $HM_{e1{\to}h1}^{n}$ represents the pseudo label generated by $Network A$ ($f_{\theta}^{A}$) for the $n$th image in the unlabeled data. 
$HM_{h1}^{n}$ represents the prediction result of $Network B$ ($f_{\theta}^{B}$) on the $n$th image output in unlabeled data.
It is worth noting that when the model is used as a teacher model, the parameters are fixed. 
When the model is used as a student model, the parameters need to be updated.

\textbf{Train Step 3.}
This step is similar to Train Step 2, except that we need to exchange the identities of $Network A$ ($f_{\theta}^{A}$) and $Network B$ ($f_{\theta}^{B}$), $Network B$ ($f_{\theta}^{B}$) as the teacher model (with fixed parameters), and $Network A$ ($f_{\theta}^{A}$) as the student model (with updated parameters).
The consistency loss is as follows:
\begin{equation}
    L_{unsup2}=\sum_{n{\in}N}{{\|{HM_{e2{\to}h2}^{n} - HM_{h2}^{n} }\|}^{2}}
\end{equation}
where $HM_{e2{\to}h2}^{n}$ represents the pseudo label generated by $Network B$ ($f_{\theta}^{B}$) for the $n$th image in the unlabeled data. 
$HM_{h2}^{n}$ represents the prediction result of $Network A$ ($f_{\theta}^{A}$) on the $n$th image output in unlabeled data.

\textbf{Train Step 4.}
We take $Network A$ ($f_{\theta}^{A}$) and $Network B$ ($f_{\theta}^{B}$) as teacher models (with fixed parameters) and $Network C$ ($f_{\theta}^{C}$) as student models (with updated parameters).
Next, we input the pseudo label $HM_{e1{\to}h1}^{n}$ and $HM_{e2{\to}h2}^{n}$ of the same image output by $Network A$ ($f_{\theta}^{A}$) and $Network B$ ($f_{\theta}^{B}$) in Train Step 3 and Train Step 4 into the PCM module.
In addition, we also input the pseudo label $HM_{e1{\to}h1}^{last, n}$ and $HM_{e2{\to}h2}^{last, n}$ generated by $Network A$ ($f_{\theta}^{A}$) and $Network B$ ($f_{\theta}^{B}$) on this image in the last epoch into the PCM module.
The PCM module is shown in Fig. \ref{fig:Fusion_method}, where $HM_{e1{\to}h1}^{n}$ corresponds to Fig. \ref{fig:Fusion_method} (b), $HM_{e2{\to}h2}^{n}$ corresponds to Fig. \ref{fig:Fusion_method} (d), $HM_{e1{\to}h1}^{last, n}$ corresponds to Fig. \ref{fig:Fusion_method} (a), and $HM_{e2{\to}h2}^{last, n}$ corresponds to Fig. \ref{fig:Fusion_method} (c).
Since the output results of the same model in two epochs may be similar, we only calculate the position inconsistency between different models.
We first post-process the generated pseudo label $HM$ to obtain pseudo keypoint coordinates.
Then, we calculate the pixel distance between different pseudo keypoints.
We normalize it with the diagonal length of the heatmap to obtain the position inconsistency:
\begin{equation}
    PI=\frac{\|argmax(HM_{i, k}^{A}) - argmax(HM_{j, k}^{B})\|}{L_{HM}}
\end{equation}
where $HM_{i, k}^{A}$ represents the pseudo label of the K$th$ keypoint output by $Network A$ ($f_{\theta}^{A}$) in the $i$th epoch, and $HM_{j, k}^{B}$ represents the pseudo label of the $K$th keypoint output by $Network B$ ($f_{\theta}^{B}$) in the $j$th epoch. 
$L_{HM}$ represents the diagonal length of the heatmap.
We select a group of pseudo labels $HM_{min1}$ and $HM_{min2}$ with the smallest position inconsistency ($PI$), and conduct pseudo label fusion to obtain the corrected pseudo labels: 
\begin{equation}
    HM_{Final} = 0.5 \cdot (HM_{min1} + HM_{min2})
\end{equation}
We use the same operation as in Train Step 2 to obtain hard samples with occlusion, and pass them into the $Network C$ ($f_{\theta}^{C}$) to get the prediction results $HM_{h3}^{n}$.
The consistency loss is as follows:
\begin{equation}
    L_{unsup3}=\sum_{n{\in}N}{{\|{HM_{Final}^{n} - HM_{h3}^{n} }\|}^{2}}
\end{equation}
The final loss function is as follows:
\begin{equation}
    L_{Final} = L_{sup} + \beta\cdot(L_{unsup1} + L_{unsup2} + L_{unsup3})
\end{equation}
where $\beta$ represents weight of unsupervised learning.

\textbf{Test.}
$Network A$  and $Network B$  are used to guide $Network C$ training. 
When testing, we will only use $Network C$. 
Therefore, our method does not increase the number of parameters or calculations of the model.

\subsection{Semi-Supervised Cut-Occlude Based on Pseudo Keypoint Perception}
\label{sec:SSCO}
\begin{figure}[t]
\centering
\includegraphics[width=0.95\linewidth]{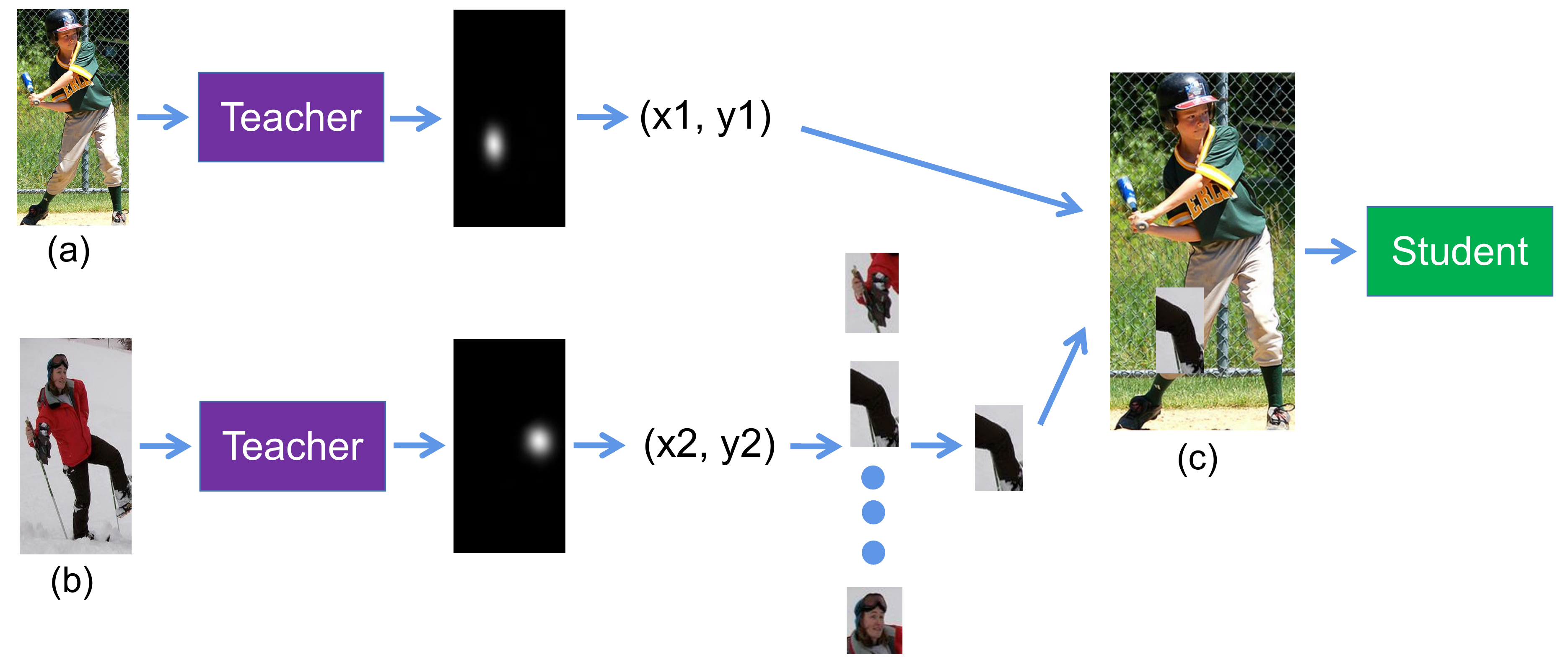}
\caption{
Semi-supervised Cut-Occlude based on pseudo keypoint perception (SSCO).
(a) and (b) are two images in one batch.
(c) is a hard sample with occlusion.
}
\label{fig:my_augmentation}
\end{figure}

One of the main difficulties in 2D HPE is occlusion. 
We use the semi-supervised Cut-Occlude based on pseudo keypoint perception to provide more hard and effective samples for student models.
Let's take two images in one batch as an example, as shown in Fig. \ref{fig:my_augmentation}.
First, we input image (a)  into the teacher model to get pseudo labels and obtain the coordinates of each pseudo keypoint through post-processing.
Next, we extract $N$ pseudo keypoint coordinates $(x1, y1)$ from them (assuming that $N$ is 1), and we take this coordinate as the center of the position to be pasted.
Then, we input image (b) into the teacher model, and we also get $N$ pseudo keypoint coordinates $(x2, y2)$, which are taken as the central coordinates of the local limb image.
We use this coordinate to clip a local limb image.
After we get the local limb image, we will paste it to the position $(x1, y1)$ in the image (a), as shown in Fig. \ref{fig:my_augmentation} (c).
Finally, we input it into the student model to get the prediction results.

\section{Experiments}
\label{sec:Experiments}

\subsection{Datasets}
\label{sec:Dataset}

\textbf{MPII \cite{andriluka14cvpr} and AI-Challenger \cite{wu2019large}.}
The MPII dataset contains 25k images and 40k person instances with 16 keypoints.
The AI-Challenger dataset has 210k images and 370K person instances with 14 keypoints.
We use MPII as the labeled set, AI-Challenger as the unlabeled set.
The metric of PCKh@0.5 \cite{andriluka14cvpr} is reported.

\textbf{COCO \cite{lin2014microsoft}.}
COCO dataset has 4 subsets: $TRAIN$, $VAL$, $TEST-DEV$ and $TEST-CHALLENGE$.
In addition, there are 123K wild unlabeled images ($WILD$).
We randomly selected 1K, 5K and 10K labeled data from $TRAIN$.
In some experiments, unlabeled data came from the remaining images of $TRAIN$.
In other experiments, we used the entire $TRAIN$ as the labeled dataset and $WILD$ as the unlabeled dataset.
The metric of mAP (Average AP over 10 OKS thresholds) \cite{lin2014microsoft} is reported.

\textbf{CEPDOF \cite{duan2020rapid}.}
This dataset is an indoor dataset collected by an overhead fisheye camera. 
It only contains bbox labels for human detection, without keypoint labels. 
We will experiment with this dataset as unlabeled data.
Since the dataset is video data, and the repeatability between adjacent frames is high, we conducted 10 times down-sampling of the original dataset, and filtered person instances whose height or width is less than 50 pixels.
Finally, there are 11878 person instances.

\begin{figure}[t]
\centering
\includegraphics[width=0.95\linewidth]{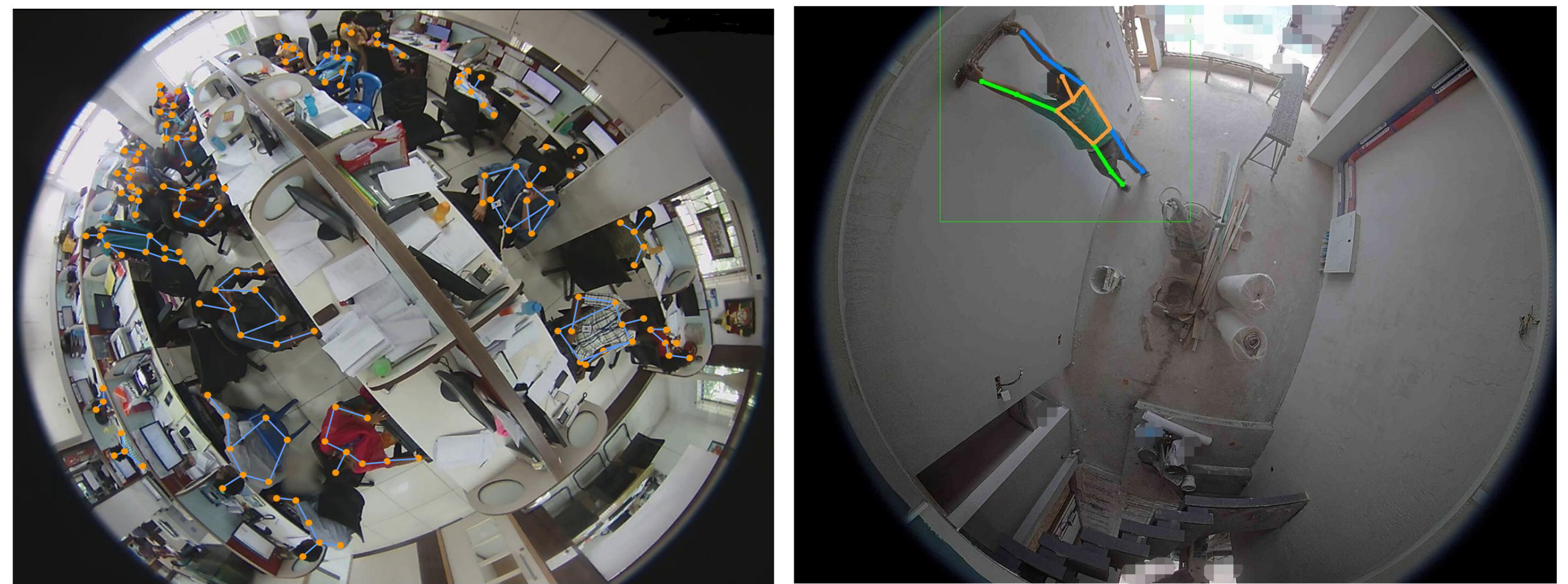}
\caption{
\textbf{Left}:  One image in the WEPDTOF-Pose dataset.
\textbf{Right}: One image in the BKFisheye dataset.
}
\label{fig:fisheye}
\end{figure}

\textbf{WEPDTOF-Pose.}
This dataset is a new human body keypoint dataset based on the WEPDTOF \cite{tezcan2022wepdtof} dataset.
We will release it soon.
It is an indoor dataset collected by an indoor overhead fisheye camera.
Since the WEPDTOF is video dataset, and the repeatability between adjacent frames is high, we conducted 10 times down-sampling of the original dataset, and filtered person instances whose height or width is less than 50 pixels.
Then, we annotate the processed images, and there are 14 keypoints in total: right shoulder, right elbow, right wrist, left shoulder, left elbow, left wrist, right hip, right knee, right ankle, left hip, left knee, left ankle, head, and lower neck, as shown in Fig. \ref{fig:fisheye} (Left).
It consists of WEPDTOF-Pose $TRAIN$ (4688 person instances) and WEPDTOF-Pose $TEST$ (1179 person instances).
The full amount of WEPDTOF-Pose $TRAIN$ is used as labeled data, and the CEPDOF \cite{duan2020rapid} dataset is used as unlabeled data for experiments.
The metric of mAP \cite{lin2014microsoft} is reported.
More details see supplementary material.

\textbf{BKFisheye.}
A dataset of a real site scene (after removing sensitive information) consists of BKFisheye $TRAIN$ (7330 person instances), BKFisheye $TEST$ (2655 person instances), and BKFisheye $UNLABEL$ (46923 person instances).
This dataset doesn't contain personal identity or other personal privacy information. 
We have mosaic the faces in the images.
The annotation method is consistent with WEPDTOF-Pose, as shown in Fig. \ref{fig:fisheye} (Right).
The metric of mAP \cite{lin2014microsoft} is reported.

\begin{table}[t]
	\centering
	\caption{AP of different methods on COCO \cite{lin2014microsoft} when different numbers of labels are used. 
	Backbone is ResNet18 \cite{he2016deep}.}
	\label{table:diffierent_number_COCO}
	\scriptsize
	\begin{tabular}{llllc}
		\toprule
		Methods     & $1$K   & $5$K   & $10$K  & All               \\
		\hline
		Supervised \cite{simplebaselines}  & 31.5  & 46.4          & 51.1         & 67.1 \\
		\hline
		PseudoPose  \cite{xie2021empirical}  & 37.2          & 50.9          & 56.0             & ---                        \\
		DataDistill \cite{radosavovic2018data}  & 37.6 & 51.6 & 56.6 & ---\\
		Cons \cite{xie2021empirical} & 42.1 & 52.3 & 57.3 & --- \\
		Dual \cite{xie2021empirical}  & 44.6 & 55.6 & 59.6 & --- \\
		\hline
		\textbf{Ours} & \textbf{46.9 \textcolor{red}{$\uparrow$15.4}} & \textbf{57.5 \textcolor{red}{$\uparrow$11.1}} & \textbf{60.7 \textcolor{red}{$\uparrow$9.6}} & --- \\
		\bottomrule                       
	\end{tabular}
\end{table}

\begin{table}[]
    \footnotesize   
	\centering
	\caption{The effects of using different network structures for the two models ($Teacher$ and $Student$) on COCO \cite{lin2014microsoft}. 
	We report the AP of the \textbf{student model}.}
	\label{table:diffmodels}

    \begin{tabular}{lccccc}
		\toprule
		Methods & Teacher & Student & 1K & 5K & 10K \\
		\hline
		
		Supervised \cite{simplebaselines} & --- & ResNet18 & 31.5 & 46.4 & 51.1 \\
		Supervised \cite{simplebaselines} & --- & ResNet50 & 34.4 & 50.3 & 56.3 \\
		\hline
		
		DUAL \cite{xie2021empirical} & ResNet18 & ResNet18 & 44.6 & 55.6 & 59.6 \\
		DUAL \cite{xie2021empirical} & ResNet50 & ResNet50 & 48.7 & 61.2 & 65.0 \\
		DUAL \cite{xie2021empirical} & ResNet50 & ResNet18 & 47.2 & 57.2 & 60.4 \\
		\hline
		
		Ours & ResNet18 & ResNet18 & 46.9 & 57.5 & 60.7 \\
		Ours & ResNet50 & ResNet50 & 49.4 & 61.6 & 65.4 \\
		Ours & ResNet50 & ResNet18 & \textbf{48.3} & \textbf{58.9}  & \textbf{61.9}  \\

    \bottomrule 
    \end{tabular}
\end{table}

\begin{table}[t]
    \centering
	\scriptsize
	\caption{Results on the COCO \cite{lin2014microsoft} $VAL$ set when all images from the $TRAIN$ set are used as the labeled set and all images from the $WILD$ set are used as the unlabeled set.}
	\label{table:largescale}
	\begin{tabular}{lclccc}
		\toprule
		Method & Network & AP & Ap .5 & AR & AR .5 \\
		\hline
		
		Supervised \cite{simplebaselines} & ResNet50  & 70.9        & 91.4            & 74.2        & 92.3  \\
		Dual \cite{xie2021empirical} & ResNet50      & 73.9   & 92.5 & 77.0 & 93.5 \\
		\textbf{Ours} & \textbf{ResNet50}   & \textbf{74.2 \textcolor{red}{$\uparrow$3.3}} & \textbf{92.7} & \textbf{77.2} & \textbf{93.8} \\
		\hline
		
		Supervised \cite{simplebaselines}  & ResNet101     & 72.5        & 92.5           & 75.6        & 93.1  \\
		Dual \cite{xie2021empirical} & ResNet101     & 75.3     & 93.6     & 78.2        & 94.1     \\
		\textbf{Ours} & \textbf{ResNet101}     & \textbf{75.5 \textcolor{red}{$\uparrow$3.0}} &  \textbf{93.8}  &  \textbf{78.4}  &  \textbf{94.2}     \\
		\hline
		
		Supervised \cite{simplebaselines} & ResNet152 &      73.2        & 92.5           & 76.3        & 93.2      \\
		Dual \cite{xie2021empirical} & ResNet152     & 75.5      & 93.6            & 78.5        & 94.3     \\
		\textbf{Ours} & \textbf{ResNet152}   & \textbf{75.7 \textcolor{red}{$\uparrow$2.5}}    & \textbf{93.7}   &  \textbf{78.6}   & \textbf{94.5}   \\
		
		\hline
		Supervised \cite{sun2019hrnet}  & HRNetW48      & 77.2        & 93.5           & 79.9        & 94.1    \\
		Dual \cite{xie2021empirical} & HRNetW48    & 79.2  & 94.6 & 81.7 & 95.1  \\
		\textbf{Ours} & \textbf{HRNetW48}    & \textbf{79.4 \textcolor{red}{$\uparrow$2.2}}  & \textbf{94.8} & \textbf{81.9} & \textbf{95.2}  \\
		\bottomrule         
	\end{tabular}
\end{table}

\subsection{Implementation Details}
Consistent with the previous work \cite{xie2021empirical}, we use SimpleBaseline to estimate the heatmap and ResNet \cite{he2016deep} and HRNet \cite{sun2019hrnet} as backbones.
The input image size is set to 256x192.
COCO \cite{lin2014microsoft} dataset training is conducted on 4 A100 GPUs, and the batch size is 32. 
We use the Adam optimizer \cite{kingma2014adam} to train these models.
The initial learning rate is 1e-3, which decreases to 1e-4 and 1e-5 at 70 epochs and 90 epochs, respectively, with a total of 100 epochs.
When using the complete COCO \cite{lin2014microsoft} dataset, the initial learning rate is 1e-3, and it drops to 1e-4 and 1e-5 at 300 epochs and 350 epochs, respectively, with a total of 400 epochs.
The fisheye dataset is trained on 1 A100 GPU, and the batch size is 32.
We use the Adam optimizer to train these models.
The initial learning rate is 1e-3, which decreases to 1e-4 and 1e-5 at 140 epochs and 180 epochs, respectively, with a total of 200 epochs.
When testing, do not flip horizontally.

\textbf{Data augmentation.}
Our data augmentation settings are consistent with previous work \cite{xie2021empirical}.
Easy data augmentation: random rotation ($-30^{\circ}$$\sim$$30^{\circ}$), random scale (0.75$\sim$1.25). 
Hard data augmentation: random rotation ($-60^{\circ}$$\sim$$60^{\circ}$), random scale (0.75$\sim$1.25). 
The random rotation range used by the fisheye dataset is ($-180^{\circ}$$\sim$$180^{\circ}$).

\begin{table}[t]
	\centering
	\caption{Comparison to the SOTA methods on the COCO \cite{lin2014microsoft} $TEST-DEV$ dataset. 
	The COCO $TRAIN$ set is the labeled set and COCO $WILD$ set is the unlabeled set. 
	The person detection results are provided by SimpleBaseline (with flipping strategy).}
	\label{table:cocotest}
	\scriptsize
\begin{tabular}{llcclccccc}
\toprule
Method & Network & \multicolumn{1}{l}{Input Size} & \multicolumn{1}{l}{AP}  & \multicolumn{1}{l}{AR} \\
    \hline
SB \cite{simplebaselines}  & ResNet50   & 256 × 192  & 70.2  & 75.8     \\
HRNet \cite{sun2019hrnet} & HRNetW48  & 384 × 288    & 75.5     & 80.5   \\

MSPN \cite{li2019rethinking}  & ResNet50   & 384 × 288   & 76.1     & 81.6   \\
DARK \cite{zhang2020distribution}  & HRNetW48   & 384 × 288   & 76.2      & 81.1   \\
UDP \cite{huang2020devil}   & HRNetW48   & 384 × 288    & 76.5     & 81.6   \\
DUAL \cite{xie2021empirical} (+DARK)    & HRNetW48    & 384 × 288   & 77.2   & 82.2  \\ 
    \hline
\textbf{Ours (+DARK) }   &  \textbf{HRNetW48 }   & \textbf{384 × 288 }  & \textbf{77.5}  &  \textbf{82.4}  \\  

\bottomrule         
\end{tabular}
\end{table}

\begin{table}[t]
	\scriptsize
	\centering
	\setlength{\tabcolsep}{3pt}
	\caption{ Comparisons on the MPII \cite{andriluka14cvpr} test set (PCKh@0.5). 
	Our method uses HRNetW32 \cite{sun2019hrnet} as backbone and size is 256 × 256. 
	The MPII and AIC (w/o labels) \cite{wu2019large} dataset are used for training.}
	\label{table:mpii_test}
    \begin{tabular}{lccccccccc}
    \toprule
    Method & Hea& Sho& Elb& Wri& Hip& Kne& Ank & Total \\
    \hline
    Newell et al. \cite{newell2016stacked}  & 98.2  & 96.3  & 91.2  & 87.1  & 90.1  & 87.4 & 83.6 & 90.9 \\
    Xiao et al. \cite{simplebaselines}   & 98.5 & 96.6 & 91.9 & 87.6 & 91.1 & 88.1 & 84.1 & 91.5 \\
    Ke et al. \cite{ke2018multi}  & 98.5  & 96.8  & 92.7  & 88.4  & 90.6  & 89.4 & 86.3 & 92.1 \\
    Sun et al. \cite{sun2019hrnet}   & 98.6 & 96.9 & 92.8 & 89.0   & 91.5 & 89.0   & 85.7 & 92.3 \\
    Zhang et al. \cite{zhang2019human}   & 98.6 & 97.0   & 92.8 & 88.8 & 91.7 & 89.8 & 86.6 & 92.5 \\
    Xie et al. \cite{xie2021empirical}    & 98.7  & 97.3  & 93.7  & 90.2  & 92.0  & 90.3 & 86.5 & 93.0  \\
    \hline
    \textbf{Ours}  & \textbf{98.7}  &  \textbf{97.5}  &  \textbf{94.0}  &  \textbf{90.6}  &  \textbf{92.5}  &  \textbf{91.1} & \textbf{87.1} & \textbf{93.3} \\ 
 
    \bottomrule  
    \end{tabular}
\end{table}

\subsection{Comparison with SOTA Methods}
Consistent with the previous work \cite{xie2021empirical}, we first use the ResNet18 \cite{he2016deep} model to conduct experiments on the COCO \cite{lin2014microsoft} dataset.
We used 1K, 5K, and 10K labeled data for the experiment, as shown in Table. \ref{table:diffierent_number_COCO}.
The results of supervised training using only labeled data are the worst.
Our method outperforms the best semi-supervised 2D human pose estimation method in 1K, 5K, and 10K settings, and improves 2.3 mAP, 1.9 mAP, and 1.1 mAP respectively.

We evaluate the effect of using different networks in Table. \ref{table:diffmodels}.
We use ResNet50 as the Teacher model and ResNet18 as the Student model.
We find that the ResNet18 model has significantly improved performance.
This is mainly because ResNet50 can provide more accurate pseudo label for ResNet18 which notably boosts its performance.
As shown in the Table. \ref{table:diffmodels}, we can use large models as teachers to improve the performance of lightweight models, and our method goes beyond the previous semi-supervised method.

Consistent with the previous work \cite{xie2021empirical}, we used the complete COCO \cite{lin2014microsoft} $TRAIN$ as the labeled dataset and $WILD$ as the unlabeled data for experiments, as shown in Table. \ref{table:largescale}.
It can be seen from Table. \ref{table:largescale} that our method is always better than the best method used in different models.
Compared with the supervised training only using the labeled part, the improvement is obvious.
We also reported our results on the COCO \cite{lin2014microsoft} $TEST-DEV$ dataset, as shown in Table. \ref{table:cocotest}.
It can be seen that although the previous methods have achieved high performance, our methods can still improve performance on this basis and outperform the previous methods \cite{
simplebaselines, sun2019hrnet, li2019rethinking, zhang2020distribution, huang2020devil, xie2021empirical}.
In addition, we also experimented with the MPII \cite{andriluka14cvpr} training set as a labeled dataset and the AIC \cite{wu2019large} dataset as an unlabeled dataset, as shown in Table. \ref{table:mpii_test}.
The results of our method on the MPII \cite{andriluka14cvpr} test set exceed those of previous methods \cite{newell2016stacked, simplebaselines, ke2018multi, sun2019hrnet, zhang2019human, xie2021empirical}.

\begin{table}[t]
	\scriptsize
	\setlength{\tabcolsep}{3pt}
	\centering
	\caption{Ablation Study.
	DUAL \cite{xie2021empirical} and JC \cite{xie2021empirical} are the previous SOTA methods. 
	PCM and SSCO are our methods.
	}
	\label{table:Ablation_study}
    \begin{tabular}{lcccc}
    \toprule
    Method                    & Backbone     & Train       & Test     & AP      \\
    \hline
    Baseline \cite{simplebaselines}  & ResNet18     & COCO 10K    & COCO     & 51.1    \\
    \hline
    +DUAL \cite{xie2021empirical}  & ResNet18     & COCO 10K    & COCO     & 58.7    \\
    +DUAL \cite{xie2021empirical} +JC \cite{xie2021empirical}   & ResNet18     & COCO 10K    & COCO     & 59.6    \\
    +DUAL \cite{xie2021empirical} +SSCO(Ours)         & ResNet18     & COCO 10K    & COCO     & 60.1    \\
    
    +PCM(Ours)              & ResNet18     & COCO 10K    & COCO     & 59.6    \\
    +PCM(Ours) +Cutout \cite{devries2017improved}         & ResNet18     & COCO 10K    & COCO     &  60.0  \\
    +PCM(Ours) +Mixup \cite{zhang2017mixup}     & ResNet18     & COCO 10K    & COCO     &   58.2  \\
    +PCM(Ours) +CutMix \cite{Yun2019cutmix}         & ResNet18     & COCO 10K    & COCO     &  60.5   \\
    +PCM(Ours) +Rand Augment \cite{cubuk2020randaugment}    & ResNet18     & COCO 10K    & COCO     &  59.9   \\
    +PCM(Ours) +JC \cite{xie2021empirical}         & ResNet18     & COCO 10K    & COCO     & 60.0    \\
    \textbf{+PCM(Ours) +SSCO(Ours)}  & \textbf{ResNet18}     & \textbf{COCO 10K}   & \textbf{COCO}   & \textbf{60.7}    \\
    \hline
    
    +DUAL \cite{xie2021empirical} +JC \cite{xie2021empirical}   & ResNet18     & COCO 5K    & COCO     & 55.6    \\
    \textbf{+PCM(Ours) +SSCO(Ours)}  & \textbf{ResNet18}     & \textbf{COCO 5K}   & \textbf{COCO}   & \textbf{57.5}    \\
    \hline
    
    +DUAL \cite{xie2021empirical} +JC \cite{xie2021empirical}   & ResNet18     & COCO 1K    & COCO     & 44.6    \\
    \textbf{+PCM(Ours) +SSCO(Ours)}  & \textbf{ResNet18}     & \textbf{COCO 1K}   & \textbf{COCO}   & \textbf{46.9}    \\
    \hline
    
    \end{tabular}
\end{table}

\begin{table}[t]
	\scriptsize
	\renewcommand\tabcolsep{4.0pt}
	\centering
	\caption{
	Hyper-parameter analysis of SSCO module. 
	$N$ represents the number of local limbs used.
	}
	\label{table:hyper-parameter}
    \begin{tabular}{lccccc}
    \toprule
    Method                &  N    & Backbone     & Train       & Test     & AP      \\
    \hline
    +PCM +SSCO    &  0    & ResNet18     & COCO 10K    & COCO     & 59.6    \\
    +PCM +SSCO           &  1    & ResNet18     & COCO 10K    & COCO     & 60.2    \\
    \textbf{+PCM +SSCO}  &  \textbf{2}    & \textbf{ResNet18}   & \textbf{COCO 10K }   & \textbf{COCO}    &  \textbf{60.7}    \\
    +PCM +SSCO           &  3    & ResNet18     & COCO 10K    & COCO     & 60.4    \\
    +PCM +SSCO           &  4    & ResNet18     & COCO 10K    & COCO     & 60.5    \\

    \hline
 
    \end{tabular}
\end{table}

\subsection{Ablation Study}
To study the effectiveness of our method, we have conducted a large number of ablation studies. 
We conducted the experiment under the setting of COCO 10K.
The model used is ResNet18 \cite{he2016deep}.

\textbf{Effect of the PCM.}
As shown in Table. \ref{table:Ablation_study}, compared with the baseline, the results obtained by using the PCM module increase 8.4 mAP.
Then, we compared the results of using the DUAL \cite{xie2021empirical} module and the PCM module alone. 
Our method is 0.8 mAP higher than the best method before.

\textbf{Effect of the SSCO.}
As shown in Table. \ref{table:Ablation_study}, by adding the SSCO module to the PCM module, 1.2 mAP can be improved.
We also added the SSCO module to the previous best method DUAL \cite{xie2021empirical}, which improved 1.4mAP.
In addition, compared with Cutout \cite{devries2017improved}, Mixup \cite{zhang2017mixup}, CutMix \cite{Yun2019cutmix}, Rand Augment \cite{cubuk2020randaugment} and JC \cite{xie2021empirical}, our SSCO shows better performance.
As shown in Table. \ref{table:hyper-parameter}, we carried out the hyper-parameter analysis of the SSCO module.
When we use 2 local limb images, we get the best results.

By observing the table. \ref{table:Ablation_study}, we can see that the performance is improved by adding \textbf{PCM}, \textbf{SSCO} modules, which verifies the effectiveness of these modules.
The less labeled data we use, the more obvious our method will be.

		
		


\begin{table}[t]
	\centering
	\caption{
	Comparison to the SOTA methods on the datasets collected by indoor overhead fisheye camera.
	WEPDTOF-Pose $TEST$ and BKFisheye $TEST$  are used as the test set.
	WEPDTOF-Pose $TRAIN$ and BKFisheye $TRAIN$ are used as the labeled set.
	CEPDOF \cite{duan2020rapid} and BKFisheye $UNLABEL$ are used as the unlabeled set.
	Backbone is ResNet18 \cite{he2016deep}.
	}
	\label{table:whole_Fisheye}
	\scriptsize
	\setlength{\tabcolsep}{1pt}
	\begin{tabular}{lccll}
		\toprule
		Methods         & Labeled Dataset    &  Unlabeled Dataset   &  AP  & AR      \\
		\hline
		Supervised \cite{simplebaselines}      &  WEPDTOF-Pose  &  ---   &  49.5    &      53.4   \\
		Cons \cite{xie2021empirical}           &  WEPDTOF-Pose  &  CEPDOF   &   54.6   &   58.1       \\
		Dual \cite{xie2021empirical}           &  WEPDTOF-Pose  &  CEPDOF   &   55.1   &    59.0     \\
		\textbf{Ours}   &  \textbf{WEPDTOF-Pose}  &  \textbf{CEPDOF}    &  \textbf{55.6 \textcolor{red}{$\uparrow$6.1}}    &  \textbf{60.0 \textcolor{red}{$\uparrow$6.6}}  \\
		\hline
		
		Supervised \cite{simplebaselines}      &  WEPDTOF-Pose  & ---   &   49.5   &   53.4      \\
		Cons \cite{xie2021empirical}           &  WEPDTOF-Pose  &  BKFisheye UNLABEL   &  57.2    &   61.4      \\
		Dual \cite{xie2021empirical}           &  WEPDTOF-Pose  &  BKFisheye UNLABEL   &   57.2   &   61.5      \\
		\textbf{Ours}   &  \textbf{WEPDTOF-Pose}  &  \textbf{BKFisheye UNLABEL}    &  \textbf{59.1 \textcolor{red}{$\uparrow$9.6}}    &  \textbf{63.7 \textcolor{red}{$\uparrow$10.3}}  \\
		\hline
		
		Supervised \cite{simplebaselines}      &  BKFisheye   &  ---  &     65.2    &    70.4    \\
		Cons \cite{xie2021empirical}            &  BKFisheye   &  BKFisheye UNLABEL  &     68.2    &    72.7    \\
		Dual \cite{xie2021empirical}            &  BKFisheye   &  BKFisheye UNLABEL  &     68.4    &  73.0     \\
		\textbf{Ours}   &  \textbf{BKFisheye}   &  \textbf{BKFisheye UNLABEL }  & \textbf{68.7 \textcolor{red}{$\uparrow$3.5}} &  \textbf{73.7 \textcolor{red}{$\uparrow$3.3}}   \\

		\bottomrule                       
	\end{tabular}
\end{table}

\subsection{Results on WEPDTOF-Pose and BKFisheye Datasets}
To further verify the effectiveness of our method, we have conducted lots of experiments on WEPDTOF-Pose and BKFisheye.
We use the complete WEPDTOF-Pose $TRAIN$ as the labeled dataset, and 11878 person instances in CEPDOF \cite{duan2020rapid} as the unlabeled dataset for experiments, as shown in Table. \ref{table:whole_Fisheye} (Top).
Then, We use the complete WEPDTOF-Pose $TRAIN$ as the labeled dataset, and BKFisheye $UNLABEL$ as the unlabeled dataset for experiments, as shown in Table. \ref{table:whole_Fisheye} (Middle).
In addition, we also conducted the same experiment on the BKFisheye dataset (labeled training set is BKFisheye $TRAIN$, unlabeled training set is BKFisheye $UNLABEL$, and the test set is BKFisheye $TEST$), as shown in Table. \ref{table:whole_Fisheye} (Down).

\section{Conclusion}
In this work, we propose a new semi-supervised 2D human pose estimation method.
We first introduce our proposed semi-supervised 2D human pose estimation framework driven by the position inconsistency pseudo label correction module.
Then, we introduce the semi-supervised Cut-Occlude based on pseudo keypoint perception.
We have carried out a lot of experiments on datasets of different scenarios, which proved the effectiveness of our method.
In addition, we released our code and new dataset, hoping to stimulate more people to study in this field.

{\small
\bibliographystyle{ieee_fullname}
\bibliography{egbib}

\begin{thebibliography}{10}\itemsep=-1pt

\bibitem{andriluka14cvpr}
Mykhaylo Andriluka, Leonid Pishchulin, Peter Gehler, and Bernt Schiele.
\newblock {2D} human pose estimation: New benchmark and state of the art
  analysis.
\newblock In {\em Proceedings of the IEEE Conference on Computer Vision and
  Pattern Recognition}, pages 3686--3693, 2014.

\bibitem{arazo2020pseudo}
Eric Arazo, Diego Ortego, Paul Albert, Noel~E O’Connor, and Kevin McGuinness.
\newblock Pseudo-labeling and confirmation bias in deep semi-supervised
  learning.
\newblock In {\em 2020 International Joint Conference on Neural Networks
  (IJCNN)}, pages 1--8. IEEE, 2020.

\bibitem{berthelot2019mixmatch}
David Berthelot, Nicholas Carlini, Ian Goodfellow, Nicolas Papernot, Avital
  Oliver, and Colin~A Raffel.
\newblock Mixmatch: A holistic approach to semi-supervised learning.
\newblock In {\em Advances in Neural Information Processing Systems}, pages
  5049--5059, 2019.

\bibitem{bin2020adversarial}
Yanrui Bin, Xuan Cao, Xinya Chen, Yanhao Ge, Ying Tai, Chengjie Wang, Jilin Li,
  Feiyue Huang, Changxin Gao, and Nong Sang.
\newblock Adversarial semantic data augmentation for human pose estimation.
\newblock In {\em European conference on computer vision}, pages 606--622.
  Springer, 2020.

\bibitem{cai2021jolo}
Jinmiao Cai, Nianjuan Jiang, Xiaoguang Han, Kui Jia, and Jiangbo Lu.
\newblock Jolo-gcn: mining joint-centered light-weight information for
  skeleton-based action recognition.
\newblock In {\em Proceedings of the IEEE/CVF Winter Conference on Applications
  of Computer Vision}, pages 2735--2744, 2021.

\bibitem{chen2018cascaded}
Yilun Chen, Zhicheng Wang, Yuxiang Peng, Zhiqiang Zhang, Gang Yu, and Jian Sun.
\newblock Cascaded pyramid network for multi-person pose estimation.
\newblock In {\em Proceedings of the IEEE conference on computer vision and
  pattern recognition}, pages 7103--7112, 2018.

\bibitem{cheng2020higherhrnet}
Bowen Cheng, Bin Xiao, Jingdong Wang, Honghui Shi, Thomas~S Huang, and Lei
  Zhang.
\newblock Higherhrnet: Scale-aware representation learning for bottom-up human
  pose estimation.
\newblock In {\em Proceedings of the IEEE/CVF conference on computer vision and
  pattern recognition}, pages 5386--5395, 2020.

\bibitem{cubuk2020randaugment}
Ekin~D Cubuk, Barret Zoph, Jonathon Shlens, and Quoc~V Le.
\newblock Randaugment: Practical automated data augmentation with a reduced
  search space.
\newblock In {\em Proceedings of the IEEE/CVF Conference on Computer Vision and
  Pattern Recognition Workshops}, pages 702--703, 2020.

\bibitem{devries2017improved}
Terrance DeVries and Graham~W Taylor.
\newblock Improved regularization of convolutional neural networks with cutout.
\newblock {\em arXiv preprint arXiv:1708.04552}, 2017.

\bibitem{duan2022revisiting}
Haodong Duan, Yue Zhao, Kai Chen, Dahua Lin, and Bo Dai.
\newblock Revisiting skeleton-based action recognition.
\newblock In {\em Proceedings of the IEEE/CVF Conference on Computer Vision and
  Pattern Recognition}, pages 2969--2978, 2022.

\bibitem{duan2020rapid}
Zhihao Duan, Ozan Tezcan, Hayato Nakamura, Prakash Ishwar, and Janusz Konrad.
\newblock Rapid: Rotation-aware people detection in overhead fisheye images.
\newblock In {\em Proceedings of the IEEE/CVF Conference on Computer Vision and
  Pattern Recognition Workshops}, pages 636--637, 2020.

\bibitem{grill2020byol}
Jean-Bastien Grill, Florian Strub, Florent Altch\'{e}, Corentin Tallec, Pierre
  Richemond, Elena Buchatskaya, Carl Doersch, Bernardo Avila~Pires, Zhaohan
  Guo, Mohammad Gheshlaghi~Azar, Bilal Piot, koray kavukcuoglu, Remi Munos, and
  Michal Valko.
\newblock Bootstrap your own latent - a new approach to self-supervised
  learning.
\newblock In H. Larochelle, M. Ranzato, R. Hadsell, M.~F. Balcan, and H. Lin,
  editors, {\em Advances in Neural Information Processing Systems}, volume~33,
  pages 21271--21284. Curran Associates, Inc., 2020.

\bibitem{he2016deep}
Kaiming He, Xiangyu Zhang, Shaoqing Ren, and Jian Sun.
\newblock Deep residual learning for image recognition.
\newblock In {\em Proceedings of the IEEE Conference on Computer Vision and
  Pattern Recognition}, pages 770--778, 2016.

\bibitem{hu2018squeeze}
Jie Hu, Li Shen, and Gang Sun.
\newblock Squeeze-and-excitation networks.
\newblock In {\em Proceedings of the IEEE conference on computer vision and
  pattern recognition}, pages 7132--7141, 2018.

\bibitem{huang2020devil}
Junjie Huang, Zheng Zhu, Feng Guo, and Guan Huang.
\newblock The devil is in the details: Delving into unbiased data processing
  for human pose estimation.
\newblock In {\em Proceedings of the IEEE/CVF Conference on Computer Vision and
  Pattern Recognition}, pages 5700--5709, 2020.

\bibitem{huang2022dh}
Linzhi Huang, Jiahao Liang, and Weihong Deng.
\newblock Dh-aug: Dh forward kinematics model driven augmentation for 3d human
  pose estimation.
\newblock {\em arXiv preprint arXiv:2207.09303}, 2022.

\bibitem{jiang2018acquisition}
Borui Jiang, Ruixuan Luo, Jiayuan Mao, Tete Xiao, and Yuning Jiang.
\newblock Acquisition of localization confidence for accurate object detection.
\newblock In {\em Proceedings of the European conference on computer vision
  (ECCV)}, pages 784--799, 2018.

\bibitem{ke2018multi}
Lipeng Ke, Ming-Ching Chang, Honggang Qi, and Siwei Lyu.
\newblock Multi-scale structure-aware network for human pose estimation.
\newblock In {\em European Conference on Computer Vision}, pages 713--728,
  2018.

\bibitem{kingma2014adam}
Diederik~P Kingma and Jimmy Ba.
\newblock Adam: A method for stochastic optimization.
\newblock In {\em International Conference on Learn-ing Representations}, 2015.

\bibitem{krizhevsky2012imagenet}
Alex Krizhevsky, Ilya Sutskever, and Geoffrey~E Hinton.
\newblock Imagenet classification with deep convolutional neural networks.
\newblock {\em Advances in neural information processing systems}, 25, 2012.

\bibitem{laine2017temporal}
Samuli Laine and Timo Aila.
\newblock Temporal ensembling for semi-supervised learning.
\newblock In {\em International Conference on Learning Representations}, 2017.

\bibitem{lee2013pseudo}
Dong-Hyun Lee.
\newblock Pseudo-label: The simple and efficient semi-supervised learning
  method for deep neural networks.
\newblock In {\em Workshop on challenges in representation learning, ICML},
  volume~3, 2013.

\bibitem{li2022pseco}
Gang Li, Xiang Li, Yujie Wang, Shanshan Zhang, Yichao Wu, and Ding Liang.
\newblock Pseco: Pseudo labeling and consistency training for semi-supervised
  object detection.
\newblock {\em arXiv preprint arXiv:2203.16317}, 2022.

\bibitem{li2021human}
Jiefeng Li, Siyuan Bian, Ailing Zeng, Can Wang, Bo Pang, Wentao Liu, and Cewu
  Lu.
\newblock Human pose regression with residual log-likelihood estimation.
\newblock In {\em Proceedings of the IEEE/CVF International Conference on
  Computer Vision}, pages 11025--11034, 2021.

\bibitem{li2022mhformer}
Wenhao Li, Hong Liu, Hao Tang, Pichao Wang, and Luc Van~Gool.
\newblock Mhformer: Multi-hypothesis transformer for 3d human pose estimation.
\newblock In {\em Proceedings of the IEEE/CVF Conference on Computer Vision and
  Pattern Recognition}, pages 13147--13156, 2022.

\bibitem{li2019rethinking}
Wenbo Li, Zhicheng Wang, Binyi Yin, Qixiang Peng, Yuming Du, Tianzi Xiao, Gang
  Yu, Hongtao Lu, Yichen Wei, and Jian Sun.
\newblock Rethinking on multi-stage networks for human pose estimation.
\newblock {\em arXiv preprint arXiv:1901.00148}, 2019.

\bibitem{li2021online}
Zheng Li, Jingwen Ye, Mingli Song, Ying Huang, and Zhigeng Pan.
\newblock Online knowledge distillation for efficient pose estimation.
\newblock In {\em Proceedings of the IEEE/CVF International Conference on
  Computer Vision}, pages 11740--11750, 2021.

\bibitem{lin2014microsoft}
Tsung-Yi Lin, Michael Maire, Serge Belongie, James Hays, Pietro Perona, Deva
  Ramanan, Piotr Doll{\'a}r, and C~Lawrence Zitnick.
\newblock Microsoft coco: Common objects in context.
\newblock In {\em European Conference on Computer Vision}, pages 740--755.
  Springer, 2014.

\bibitem{martinez2017simple}
Julieta Martinez, Rayat Hossain, Javier Romero, and James~J Little.
\newblock A simple yet effective baseline for 3d human pose estimation.
\newblock In {\em Proceedings of the IEEE international conference on computer
  vision}, pages 2640--2649, 2017.

\bibitem{newell2017associative}
Alejandro Newell, Zhiao Huang, and Jia Deng.
\newblock Associative embedding: End-to-end learning for joint detection and
  grouping.
\newblock {\em Advances in neural information processing systems}, 30, 2017.

\bibitem{newell2016stacked}
Alejandro Newell, Kaiyu Yang, and Jia Deng.
\newblock Stacked hourglass networks for human pose estimation.
\newblock In {\em European Conference on Computer Vision}, pages 483--499,
  2016.

\bibitem{pavllo20193d}
Dario Pavllo, Christoph Feichtenhofer, David Grangier, and Michael Auli.
\newblock 3d human pose estimation in video with temporal convolutions and
  semi-supervised training.
\newblock In {\em Proceedings of the IEEE/CVF Conference on Computer Vision and
  Pattern Recognition}, pages 7753--7762, 2019.

\bibitem{radosavovic2018data}
Ilija Radosavovic, Piotr Doll{\'a}r, Ross Girshick, Georgia Gkioxari, and
  Kaiming He.
\newblock Data distillation: Towards omni-supervised learning.
\newblock In {\em Proceedings of the IEEE conference on computer vision and
  pattern recognition}, pages 4119--4128, 2018.

\bibitem{rizve2021defense}
Mamshad~Nayeem Rizve, Kevin Duarte, Yogesh~S Rawat, and Mubarak Shah.
\newblock In defense of pseudo-labeling: An uncertainty-aware pseudo-label
  selection framework for semi-supervised learning.
\newblock {\em arXiv preprint arXiv:2101.06329}, 2021.

\bibitem{sajjadi2016ts}
Mehdi Sajjadi, Mehran Javanmardi, and Tolga Tasdizen.
\newblock Regularization with stochastic transformations and perturbations for
  deep semi-supervised learning.
\newblock In {\em Proceedings of the 30th International Conference on Neural
  Information Processing Systems}, page 1171–1179, Red Hook, NY, USA, 2016.
  Curran Associates Inc.

\bibitem{simonyan2014very}
Karen Simonyan and Andrew Zisserman.
\newblock Very deep convolutional networks for large-scale image recognition.
\newblock {\em arXiv preprint arXiv:1409.1556}, 2014.

\bibitem{sohn2020fixmatch}
Kihyuk Sohn, David Berthelot, Chun-Liang Li, Zizhao Zhang, Nicholas Carlini,
  Ekin~D Cubuk, Alex Kurakin, Han Zhang, and Colin Raffel.
\newblock Fixmatch: Simplifying semi-supervised learning with consistency and
  confidence.
\newblock In {\em Neural Information Processing Systems}, 2020.

\bibitem{sun2019hrnet}
Ke Sun, Bin Xiao, Dong Liu, and Jingdong Wang.
\newblock Deep high-resolution representation learning for human pose
  estimation.
\newblock In {\em Proceedings of the IEEE conference on computer vision and
  pattern recognition}, pages 5693--5703, 2019.

\bibitem{szegedy2015going}
Christian Szegedy, Wei Liu, Yangqing Jia, Pierre Sermanet, Scott Reed, Dragomir
  Anguelov, Dumitru Erhan, Vincent Vanhoucke, and Andrew Rabinovich.
\newblock Going deeper with convolutions.
\newblock In {\em Proceedings of the IEEE conference on computer vision and
  pattern recognition}, pages 1--9, 2015.

\bibitem{tarvainen2017mean}
Antti Tarvainen and Harri Valpola.
\newblock Mean teachers are better role models: Weight-averaged consistency
  targets improve semi-supervised deep learning results.
\newblock In {\em Advances in neural information processing systems}, pages
  1195--1204, 2017.

\bibitem{tezcan2022wepdtof}
Ozan Tezcan, Zhihao Duan, Mertcan Cokbas, Prakash Ishwar, and Janusz Konrad.
\newblock Wepdtof: A dataset and benchmark algorithms for in-the-wild people
  detection and tracking from overhead fisheye cameras.
\newblock In {\em Proceedings of the IEEE/CVF Winter Conference on Applications
  of Computer Vision}, pages 503--512, 2022.

\bibitem{wang2021data}
Zhenyu Wang, Yali Li, Ye Guo, Lu Fang, and Shengjin Wang.
\newblock Data-uncertainty guided multi-phase learning for semi-supervised
  object detection.
\newblock In {\em Proceedings of the IEEE/CVF Conference on Computer Vision and
  Pattern Recognition}, pages 4568--4577, 2021.

\bibitem{wu2019large}
J. {Wu}, H. {Zheng}, B. {Zhao}, Y. {Li}, B. {Yan}, R. {Liang}, W. {Wang}, S.
  {Zhou}, G. {Lin}, Y. {Fu}, Y. {Wang}, and Y. {Wang}.
\newblock Large-scale datasets for going deeper in image understanding.
\newblock In {\em IEEE International Conference on Multimedia and Expo (ICME)},
  pages 1480--1485, 2019.

\bibitem{simplebaselines}
Bin Xiao, Haiping Wu, and Yichen Wei.
\newblock Simple baselines for human pose estimation and tracking.
\newblock In {\em European Conference on Computer Vision}, pages 466--481,
  2018.

\bibitem{xie2020self}
Qizhe Xie, Minh-Thang Luong, Eduard Hovy, and Quoc~V Le.
\newblock Self-training with noisy student improves imagenet classification.
\newblock In {\em Proceedings of the IEEE/CVF Conference on Computer Vision and
  Pattern Recognition}, pages 10687--10698, 2020.

\bibitem{xie2021empirical}
Rongchang Xie, Chunyu Wang, Wenjun Zeng, and Yizhou Wang.
\newblock An empirical study of the collapsing problem in semi-supervised 2d
  human pose estimation.
\newblock In {\em Proceedings of the IEEE/CVF International Conference on
  Computer Vision}, pages 11240--11249, 2021.

\bibitem{yan2018spatial}
Sijie Yan, Yuanjun Xiong, and Dahua Lin.
\newblock Spatial temporal graph convolutional networks for skeleton-based
  action recognition.
\newblock In {\em Thirty-second AAAI conference on artificial intelligence},
  2018.

\bibitem{yang2022mix}
Lei Yang, Xinyu Zhang, Li Wang, Minghan Zhu, Chuang Zhang, and Jun Li.
\newblock Mix-teaching: A simple, unified and effective semi-supervised
  learning framework for monocular 3d object detection.
\newblock {\em arXiv preprint arXiv:2207.04448}, 2022.

\bibitem{yarowsky1995unsupervised}
David Yarowsky.
\newblock Unsupervised word sense disambiguation rivaling supervised methods.
\newblock In {\em 33rd annual meeting of the association for computational
  linguistics}, pages 189--196, 1995.

\bibitem{Yun2019cutmix}
Sangdoo Yun, Dongyoon Han, Sanghyuk Chun, Seong~Joon Oh, Youngjoon Yoo, and
  Junsuk Choe.
\newblock Cutmix: Regularization strategy to train strong classifiers with
  localizable features.
\newblock In {\em 2019 IEEE/CVF International Conference on Computer Vision
  (ICCV)}, pages 6022--6031, 2019.

\bibitem{zhang2020distribution}
Feng Zhang, Xiatian Zhu, Hanbin Dai, Mao Ye, and Ce Zhu.
\newblock Distribution-aware coordinate representation for human pose
  estimation.
\newblock In {\em Proceedings of the IEEE/CVF conference on computer vision and
  pattern recognition}, pages 7093--7102, 2020.

\bibitem{zhang2019fast}
Feng Zhang, Xiatian Zhu, and Mao Ye.
\newblock Fast human pose estimation.
\newblock In {\em Proceedings of the IEEE/CVF Conference on Computer Vision and
  Pattern Recognition}, pages 3517--3526, 2019.

\bibitem{zhang2017mixup}
Hongyi Zhang, Moustapha Cisse, Yann~N Dauphin, and David Lopez-Paz.
\newblock mixup: Beyond empirical risk minimization.
\newblock {\em arXiv preprint arXiv:1710.09412}, 2017.

\bibitem{zhang2019human}
Hong Zhang, Hao Ouyang, Shu Liu, Xiaojuan Qi, Xiaoyong Shen, Ruigang Yang, and
  Jiaya Jia.
\newblock Human pose estimation with spatial contextual information.
\newblock {\em arXiv preprint arXiv:1901.01760}, 2019.

\bibitem{zheng20213d}
Ce Zheng, Sijie Zhu, Matias Mendieta, Taojiannan Yang, Chen Chen, and Zhengming
  Ding.
\newblock 3d human pose estimation with spatial and temporal transformers.
\newblock In {\em Proceedings of the IEEE/CVF International Conference on
  Computer Vision}, pages 11656--11665, 2021.

\end{thebibliography}
}

\end{document}